\documentclass[letterpaper]{article} 
\usepackage{aaai23}  
\usepackage{times}  
\usepackage{helvet}  
\usepackage{courier}  
\usepackage[hyphens]{url}  
\usepackage{graphicx} 
\usepackage{natbib}  
\usepackage{caption} 
\usepackage{algorithmic}
\usepackage[ruled]{algorithm2e}
\usepackage{color}
\usepackage{xspace}
\usepackage{xcolor}
\usepackage{tikz}
\usepackage{comment}
\usepackage{mathrsfs}
\usepackage{amsmath}
\usepackage{diagbox}
\usepackage{booktabs}
\usepackage{enumitem}
\usepackage{multirow}
\usepackage{multicol}
\usepackage{bm}
\usepackage{amsfonts}
\usepackage{cancel}
\definecolor{citecolor}{RGB}{0, 113, 188}
\definecolor{greencolor}{RGB}{34, 139, 34}

\newcommand*{\eg}{\emph{e.g.}\@\xspace}

\newcommand*{\ie}{\emph{i.e.}\@\xspace}

\usepackage{hyperref}
\hypersetup{breaklinks=true,colorlinks,citecolor=citecolor}

\usepackage{newfloat}

\title{Deeper Insights into the Robustness of ViTs towards Common Corruptions}

\newcommand{\Drop}[1]{\textcolor{greencolor}{\small{(\bf $\downarrow$#1)}}}
\newcommand{\Rise}[1]{\textcolor{citecolor}{\small{(\bf $\uparrow$#1)}}}

\usepackage{bibentry}

\begin{document}

\author{Rui Tian$^{1,2}$ \quad Zuxuan Wu$^{1}$ \quad Qi Dai$^{2}$ \quad Han Hu$^{2}$ \quad Yu-Gang Jiang$^{1}$\\
\vspace{0.1in}
\normalsize\textmd{$^{1}$Shanghai Key Lab of Intelligent Information Processing, \\ \normalsize School of Computer Science, Fudan University\\
\normalsize$^{2}$Microsoft Research Asia}
}

\maketitle

\begin{abstract}
With Vision Transformers (ViTs) making great advances in a variety of computer vision tasks, recent literature have proposed various variants of vanilla ViTs to achieve better efficiency and efficacy. However, it remains unclear how their unique architectures impact robustness towards common corruptions. In this paper, we make the first attempt to probe into the robustness gap among ViT variants and explore underlying designs that are essential for robustness. Through an extensive and rigorous benchmarking, we demonstrate that simple architecture designs such as overlapping patch embedding and convolutional feed-forward network (FFN) can promote the robustness of ViTs. Moreover, since training ViTs relies heavily on data augmentation, whether previous CNN-based augmentation strategies that are targeted at robustness purposes can still be useful is worth investigating. We explore different data augmentation on ViTs and verify that adversarial noise training is powerful while fourier-domain augmentation is inferior. Based on these findings, we introduce a novel conditional method of generating dynamic augmentation parameters conditioned on input images, offering state-of-the-art robustness towards common corruptions.
\end{abstract}

\section{Introduction}
The Vision Transformer (ViT) \cite{dosovitskiy2021an} has propelled overwhelming advances in computer vision. 
Meanwhile, the brittleness of deep neural networks motivates extensive studies exploring the robustness of models. Recently, datasets benchmarking robustness towards common corruptions~\cite{hendrycks2018benchmarking, kar20223d} have been introduced, attracting a growing interest in tackling out-of-distribution corruptions.

To further boost efficacy and efficiency, there have emerged numerous ViT-fashion models varying in architecture \cite{liu2021swin,hassani2021escaping,yuan2021tokens,Mahmood_2021_ICCV}. Whereas previous studies have unveiled the superiority of ViTs over CNNs in robustness \cite{bai2021transformers,Mahmood_2021_ICCV,naseer2021intriguing,paul2021vision} and uncovered that characteristic self-attention mechanisms of ViTs may boost their resistance towards corruptions \cite{bhojanapalli2021understanding,paul2021vision}, most emphasis is laid only on vanilla ViTs and no light is shed on robustness gaps among ViT backbone variants. Therefore, we undertake the first exploration by reviewing  popular ViT backbones. Inspired by the improved robustness from PVTv1s~\cite{herrmann2021pyramid} to PVTv2s~\cite{wang2021pvtv2}, we unveil that even minor design strategies may conduce to robustness of ViTs, \ie overlapping patch embeddings and convolutions in feed-forward network (FFN).

In recent literature, plentiful effective data augmentation methods have been proposed to tackle robustness threats on CNNs \cite{hendrycks2021many,HendrycksMCZGL20,modas2021prime}. In additon, the training of ViTs  heavily relies on data augmentation,  yet there lacks work relating augmentation strategies with the robustness of ViTs. To provide a new insight, we explore the feasibility of implanting previous CNN-based augmentation on ViTs for better robustness. Since Mixup \cite{zhang2017mixup}, CutMix \cite{yun2019cutmix} and AutoAugment \cite{cubuk2018autoaugment} have already been ingrained in the training of ViTs, we lay our main emphasis on more complex methods, \ie, adversarial noise training (ANT) \cite{rusak2020simple} and augmentation in the frequency domain \cite{chen2021amplitude}. Furthermore, motivated by the success of dynamic networks, we introduce a simple approach that improves ANT by integrating the idea of conditional computation so as to dynamically generate sample-specific augmentation parameters.

In this paper, we aim to investigate the robustness influence of backbone designs and augmentation methods. In particular, we leverage popular Vision Transformers for evaluation and systematically study the robustness gap among ViT variants and the effectiveness of architectural designs. To be specific, we experiment on top of DeiT~\cite{touvron2021training} and Swin Transformer~\cite{liu2021swin} and validate that overlapping patch embeddings, which help to capture more local continuity information, can effectively 
lead to drop in mCE of ImageNet-C on DeiT-S by 2.2 and 5.85 for Swin-T. In addition, adding depth-wise convolution to feed-forward networks not only models better spatial relationships but also leads to  decrease of ImageNet-C mCE on DeiT-S and Swin-T by 2.19 and 6.68 respectively. By combining two designs, we can see a decline of mCE on DeiT-S and Swin-T by 2.41 and 7.86 as well as increase of 1.29 for DeiT-S and 2.56 for Swin-T on ImageNet-3DCC accuracy.

We also explore two CNN-based augmentation approaches on ViTs and discover that the training DeiT-S with adversarial noise training (ANT)~\cite{rusak2020simple} is able to boost its Top-1 accuracy on ImageNet-C from 57.20\% to 62.74\%. Furthermore, our proposed conditional augmentation which combines adversarial noise training with dynamic networks and achieves state-of-the-art mCE on ImageNet-C of 40.68 with DeiT-B and accuracy on ImageNet-3DCC of 65.31\%. 

Conclusively, we derive meaningful findings from an extensive study on Vision Transformers. Our contributions can be summarized as:
\begin{itemize}
 \setlength\itemsep{0.05in}
\item We demonstrate that simple design choices, \ie overlapping patch embeddings and convolutional feed-forward networks, can improve the resilience of ViTs towards common corruptions.

\item ViTs rely on fundamental augmentation techniques (\eg Mixup, CutMix) in training to a great extent. We verify that adversarial noise training also exhibits advantages on ViTs and structural noise serves as a stronger incentive to promote robustness. On the other hand, we show amplitude-phase interpolation fails to produce gains on ViTs. 

\item We introduce a dynamic augmentation strategy which is built upon adversarial noise training and conditional convolution. This simple yet effective approach achieves state-of-the-art robustness performance on ImageNet-C and ImageNet-3DCC. 

\end{itemize}

\section{Related Work} \label{sec:sec2}

\noindent \textbf{Common corruptions and perturbations} In the wake of ImageNet-C~\cite{hendrycks2018benchmarking} and ImageNet-3DCC~\cite{kar20223d}, which benchmark robustness towards safety-critical real-world corruptions, extensive studies have committed to overcoming the vulnerability of deep networks towards common corruptions~\cite{chen2021amplitude,hendrycks2021many,hendrycks2019augmix,HendrycksMCZGL20,hendrycks2021pixmix,li2021feature,modas2021prime}. DeepAugment \cite{hendrycks2021many} distorts images by perturbing image-to-image networks. MoEx~\cite{li2021feature} increases stability of models through swapping mean and standard deviation of image latent features. PixMix \cite{hendrycks2021pixmix} uses fractals to create images with structural complexity. Augmentation involve automatic learning is also burgeoning \cite{cubuk2018autoaugment,cubuk2020randaugment,HendrycksMCZGL20,zhang2019adversarial}. AutoAugment \cite{cubuk2018autoaugment} assigns image with best augmentation policy found by searching algorithm. AugMix \cite{HendrycksMCZGL20} augments images with stochastic and diverse augmentation methods controlled by Jensen-Shannon Divergence. TeachAugment \cite{suzuki2022teachaugment} introduces a teacher network to optimize the search space of adversarial augmentation. In this paper, we focus on robustness of ViTs towards common corruptions by exploring architecture designs and revisiting previous augmentation-based methods.


\vspace{0.05in}
\noindent \textbf{Robustness of ViTs}
Recent studies have reported ViTs' encouraging characteristic of robustness \cite{bai2021transformers,naseer2021intriguing,paul2021vision}, suggesting that the self-attention architectures contribute to their advantages over CNNs on generalization \cite{bai2021transformers} and demystifying that ViTs tend to be less biased to texture \cite{naseer2021intriguing}.
Effort has also been made to further enhance robustness on ViTs. By improving on Transformer-based components, Robust Vision Transformer (RVT) is designed to boost performance of model robustness and generalization~\cite{mao2021towards}. Mao et al. build robust global feature representations at the cost of using pre-training to generate discrete tokens to combine with continuous pixel tokens, however making the framework less handy~\cite{mao2022discrete}. AdvProp  
is beneficial on CNNs with carefully-designed adversarial training~\cite{xie2020adversarial}, and ViTs can be hardened remarkably through refining attack with pyramidal setting \cite{herrmann2021pyramid}. However, the required multi-step attack demands extensive amount of computing resources. Whereas above-mentioned works have probed into ViTs robustness towards common corruptions, gap between different ViT variants is ignored. In this paper, we make attempts to explore architecture designs beneficial to ViTs robustness. Moreover, we adopt relatively simple yet strong plug-and-play augmentation strategies to boost robustness of ViTs to their full potential.

\vspace{0.05in}
\noindent \textbf{Adversarial augmentation}
Adversarial training is widely acknowledged as strategy of defending models from attacks. In general, it executes attack and consequent defense alternately on the basis of a min-max optimization. Motivated by the effectiveness of adversarial training, min-max optimization is tailored to a broader variety of tasks, including coping with common corruptions risks \cite{herrmann2021pyramid,shu2021adversarial,xie2020adversarial}. Mild and much less time-consuming data augmentation can serve as the attack function as well as bring about considerable improvements against common corruptions \cite{shu2021adversarial}. Adversarial Noise Training (ANT) \cite{rusak2020simple} plays a beneficial role 
by adding random gaussian noise adjusted by a simple 4-layer convolution trained in adversarial manner. AugMax \cite{wang2021augmax} achieves significant performance rise by learning a worst-case combination of random augmentation. Adversarial Batch Normalization (AdvBN) \cite{shu2021encoding} withstands corruptions through generating the most difficult perturbations on mean and standard deviation of features. To our knowledge, any differentiable data augmentation can be practiced adversarially. Hence, we take advantage of different adversarial data augmentation in our conditional augmentation for improved results. 

\vspace{0.05in}
\noindent \textbf{Dynamic networks}
Through adapting parameters and structures to specific input, neural networks are transformed into dynamic states, contributing gains in accuracy \cite{chen2020dynamic,yang2019condconv}, efficiency \cite{rao2021dynamicvit,sharma2018classification} or adaptability \cite{zhu2021dynamic}. According to \cite{han2021dynamic}, dynamic networks can be divided into three different categories focusing respectively on input samples, spatial locations and temporal information. Regarding with instance-wise dynamic networks, convolutions with dynamic parameters \cite{chen2020dynamic,yang2019condconv,zhang2020dynet} bring about increase in capacity of representations with high computational efficiency and cope with input better with feeding selected informative features into models \cite{sharma2018classification}. While extensive data augmentation methods have been designed, most existing algorithms are implemented identically for different samples. By virtue of dynamic networks, we propose a novel conditional augmentation strategy by generating input-dependent dynamic augmentation parameters.

\section{Overview}
Vision Transformers have achieved remarkable results due to their high capacity in modeling global relationships among tokenized patches. 
To reduce training cost and capture local spatial information, researchers resort to various backbone designs and training phillopsohies. 
While these approaches have shown increased efficacy and efficiency, their impact on robustness of Vision Transformers yet remain unexplored. This motivates us to investigate underlying architecture designs and augmentation strategies to improve the robustness of ViTs towards common corruptions. Towards this end, we begin by introducing the experimental settings, and then we elaborate different architectural and augmentation strategies that could potentially influence the robustness of ViTs.

\subsection{Training Vision Transformers}
We conduct training following similar procedure with DeiT~\cite{touvron2021training} by using Adam-W~\cite{loshchilov2018decoupled} optimizer with a cosine decay learning rate scheduler. When training from scratch, ViTs are trained for 300 epochs with learning rate set to $1e^{-3}$ and batch size of 1024. For experiments requires fine-tuning, we decrease the learning rate to $4e^{-4}$. During training, same augmentation and regularization strategies with DeiT are kept unless mentioned otherwise.

\subsection{Robustness Datasets}
Basically, we train and validate ViTs on ImageNet and test their robustness performance on benchmarking dataset ImageNet-C~\cite{hendrycks2018benchmarking}, which provides comprehensive common corruptions in 2D setting. Also, we report results on ImageNet-3DCC~\cite{kar20223d}, which includes challenging 3D common corruptions. 

\begin{table}[htbp]
  \centering
  \caption{Results of different ViT backbones on ImageNet-C and ImageNet-3DCC. The listed models are sorted by parameters in ascending order. For simplicity, only representative ViTs are demonstrated. We append extended results in supplementary material.}
  \label{tab:vit_eval}
  \resizebox{\linewidth}{!}{\begin{tabular}{lccc}
    \toprule
    \multirow{2}{*}{\textbf{Model}} & {\textbf{IN}} & {\textbf{IN-C}} & {\textbf{IN-3DCC}} \\
    ~ & \textbf{Acc} $\uparrow$ &\textbf{mCE}$\downarrow$ & \textbf{Acc}$\uparrow$ \\
    \cmidrule{1-1} \cmidrule{2-4} 
    PiT-T~\cite{heo2021rethinking} & 72.84 & 69.11 & 47.98\\
    DeiT-T~\cite{touvron2021training} & 72.14 & 71.13 & 47.44\\
    PVTv1-T~\cite{wang2021pyramid} & 75.00 &  \textbf{79.56} &  \textbf{46.28} \\
    PVTv2-B1~\cite{wang2021pvtv2}  & 78.70 & 62.65 & 53.27\\ 
    \cmidrule{1-1} \cmidrule{2-4} 
    DeiT-S~\cite{touvron2021training} & 79.83 & 54.60 & 57.60 \\
    PiT-S~\cite{heo2021rethinking} & 80.98 & 52.47 & 58.27 \\
    PVTv1-S~\cite{wang2021pyramid} & 79.79 & \textbf{66.89} & \textbf{53.94}\\
    PVTv2-B2~\cite{wang2021pvtv2} & 82.02 & 52.56 & 59.06\\
    Swin-T~\cite{liu2021swin} & 81.16 & \textbf{61.96} & \textbf{55.90}\\
    \cmidrule{1-1} \cmidrule{2-4} 
    PVTv1-M~\cite{wang2021pyramid} & 81.31 & \textbf{62.39} &  \textbf{56.49}\\
    Swin-S~\cite{liu2021swin} & 83.17 & 54.92 & 59.56\\
    PVTv1-L~\cite{wang2021pyramid} & 81.72 & \textbf{59.86} & \textbf{57.80}\\
    PiT-B~\cite{heo2021rethinking} & 82.39 & 48.16 & 61.21\\
    PVTv2-B5~\cite{wang2021pvtv2} & 83.77 & 45.90 & 62.96\\
    DeiT-B~\cite{touvron2021training} & 81.80 & 48.52 & 61.45 \\
    Swin-B~\cite{liu2021swin} & 83.42 & \textbf{54.45} & \textbf{59.93} \\
    \bottomrule
\end{tabular}}
\end{table}

\section{Architecture Designs}

Vision Transformers capture global features in virtue of convolution-free self-attention, which is also believed to contribute to better  robustness~\cite{bhojanapalli2021understanding,paul2021vision}.  While a variety of ViT-based backbones have been introduced recently in order to achieve improved recognition results or efficiency, whether they possess better robustness remains unknown. 

In consequence, we evaluate prevailing backbones over ImageNet-C and ImageNet-3DCC. As shown in Table \ref{tab:vit_eval}, the gap between different ViT variants should NOT be ignored. Especially, compared with DeiT~\cite{touvron2021training} and PiT~\cite{heo2021rethinking}, PVTv1~\cite{herrmann2021pyramid} and Swin Transformer~\cite{liu2021swin} suffer acute weakness on corrupted data (bold in Table).

Furthermore, it's worth mentioning that growth between PVTv2s and PVTv1s on ImageNet-C and ImageNet-3DCC is considerbale. As Table \ref{tab:vit_eval} displays, PVTv2-B5 outweighs PVTv1-L on ImageNet-C by 13.96 in mCE and is 5.16\% higher in ImageNet-3DCC Top-1 accuracy. Inspired by the encouraging improvement, we start by investigating difference between these two backbones. On top of PVTv1s, PVTv2s incorporate two minor modifications into the backbone architecture, including (1) \emph{overlapping patch embedding} and (2) \emph{convolutional feed-forward networks}. To further verify their effectiveness on robustness towards common corruptions, we apply similar modifications to DeiT-S and Swin-Tiny respectively .

\begin{figure}[b]
\centering
\includegraphics[width=\linewidth]{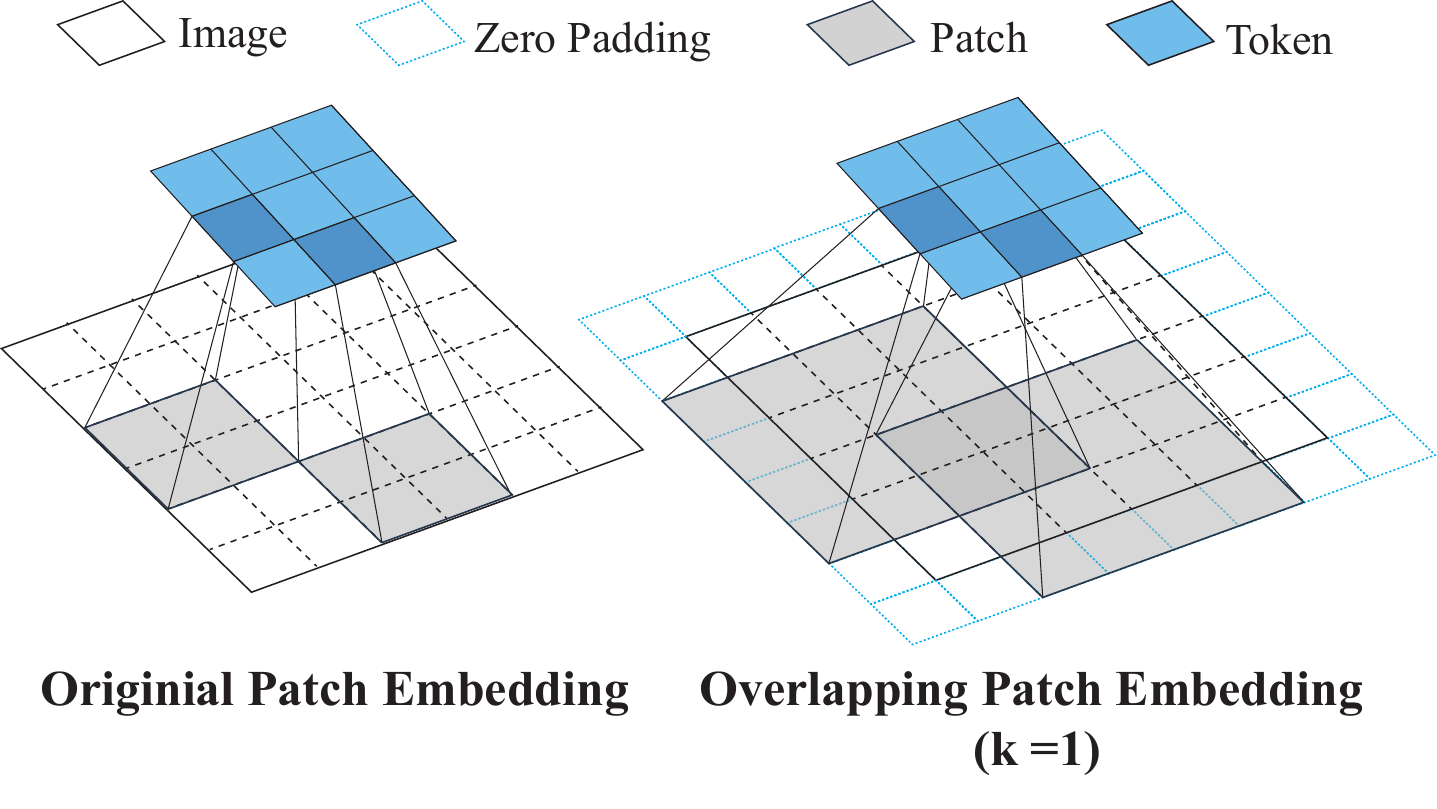}
\caption{Comparison between original patch embedding and overlapping patch embedding.}
\label{fig:ope_cmp}
\end{figure}

\subsection{Overlapping Patch Embedding} 
In vanilla ViTs~\cite{dosovitskiy2021an}, encoding of an image $x \in \mathbb{R}^{C \times H \times W}$ starts with translating it into a sequence of $T$ patches sized of $p \times p$. Flattened patches $x_p \in \mathbb{R}^{T \times (p^2 \times C)}$ are subsequently mapped to tokens $t \in \mathbb{R}^{T \times D}$ by a fully-connected layer. In DeiT \cite{touvron2021training}, patch embedding is instead instantiated with convolution whose kernel size and stride are $p$. As a result, pixels within a single patch share the identical kernel. 

In contrast to these non-overlapping patch embeddings, CNNs apply convolutions on overlapping spatial areas. Drawing inspiration from such a design, overlapping patch embedding is introduced by expanding patch window to surrounding pixels of original patch. To encode an extended area within $k$ pixels around the original patch for each single token, we perform convolution with a kernel of $((p + 2k) \times (p + 2k))$ and leave the stride size fixed by adding zero padding . Hence, overlapping pixels are encoded by neighbouring tokens while the output number of tokens is kept unchanged, as depicted in Figure \ref{fig:ope_cmp}.

\begin{table}[htbp]
  \centering
  \caption{Results of ViTs with different patch embedding designs. OPE. stands for overlapping patch embedding. When $k = 0$, models are identical with original designs.}
{\begin{tabular}{ll|ccc}
        \toprule
        \multirow{2}{*}{\textbf{Model}} & {\textbf{OPE.}} & {\textbf{IN}} & \multicolumn{2}{c}{\textbf{IN-C}} \\
        ~ & {\textbf{$k$}} & {\textbf{Acc}$\uparrow$} &  {\textbf{Acc}$\uparrow$} &  {\textbf{mCE}$\downarrow$}\\
        \cmidrule{1-1} \cmidrule{2-5} 
        \multirow{4}{*}{DeiT-S} & 0 & 79.83 & 57.20 & 54.60 \\
        ~ & 1 & 80.45 & \textbf{58.93} 
        & \textbf{52.40} \\
        ~ & 2 & 80.40 & 58.37 & 53.10 \\
        ~ & 4 & 79.77 & 58.51 & 52.94 \\
        \midrule
        \multirow{4}{*}{Swin-T} & 0 & 81.16 & 51.59 & 61.96 \\
        ~ & 1 &81.61 & 54.75 & 57.74 \\
        ~ & 2 & 81.59 & 54.96 & 57.54 \\
        ~ & 4 & 81.52 & \textbf{56.11} 
        & \textbf{56.11} \\
    \bottomrule
    \end{tabular}}
\label{tab:patch_eval}
\end{table}

In practice, with $p$ set to 16, we conduct experiments with $k$ varying in 1, 2 and 4 on DeiT and Swin Transformer. Results from Table \ref{tab:patch_eval} display a convincing conclusion that \textit{{overlapping patch embeddings conduce to robustness}}. DeiT-S drops by \textbf{2.2} in mCE when $k=1$ while Swin-T with $k = 4$ decreases \textbf{5.96} in mCE. Specifically, DeiT-S with overlapping patch embedding presents more stability towards perturbations of noise, contrast and JPEG compression (Please refer to supplementary material for details).

\subsection{Convolutional Feed-Forward Network} 

Researchers have been trying to couple ViT blocks with convolutions either to unleash efficiency of ViTs \cite{wu2021cvt,yuan2021incorporating} or to improve encoding of spatial information~\cite{islam2019much,chu2021conditional}. Depth-wise convolution (DWConv.) \cite{chollet2017xception} is a special form of convolution which applies a single filter to each input channel (input depth). We follow PVTv2 by adding a depth-wise convolution to each feed-forward block in DeiT. We denote $f_i^{1} = [f_{i, cls}^{1}, f_{i}^{1*}]$  as the output of $i$-th attention layer,  $f_{i, cls}^{1}$ represents the class token and $f_{i}^{1*}$ is further transformed with depth-wise convolution as follows:

\begin{table}[t]
  \centering
  \caption{Results of models with different feed-forward designs. When choosing w/o Conv., models are identical with original setting. We set $N$ to 32 and 4 for conditional DWConv for DeiT-S and Swin-T respectively.}
  \vspace{0.05in}
  \label{tab:ffn_eval}
    \normalsize
  {\begin{tabular}{l l|ccc}
        \toprule
        \multirow{2}{*}{\textbf{Model}}  & {\textbf{Feed-Forward}} & {\textbf{IN}} & \multicolumn{2}{c}{\textbf{IN-C}} \\
        ~ & {\textbf{Design}} & {\textbf{Acc}$\uparrow$} &  {\textbf{Acc}$\uparrow$} &  {\textbf{mCE}$\downarrow$}\\
        \cmidrule{1-1} \cmidrule{2-5} 
    \multirow{3}{*}{DeiT-S} & w/o Conv. & 79.83 & 57.20 & 54.60 \\
    ~ & DWConv. & 79.93 & \textbf{58.91} & \textbf{52.41} \\
    ~ & Cond DWConv. & 79.79 & 58.27 & 53.27\\
    \midrule
    \multirow{3}{*}{Swin-T} & w/o Conv. & 81.16 & 51.59 & 61.96 \\
    ~ & DWConv.& 81.83 & \textbf{56.73} & \textbf{55.28} \\
    ~ & Cond DWConv.& 81.95 & 56.69 & 55.34 \\

\bottomrule
\end{tabular}}
\end{table}

\begin{equation}
\begin{aligned}
f_i^{2*} &= \texttt{FC}_i^{2}(\texttt{GeLU}(\texttt{DWConv}_i(\texttt{FC}_i^{1}(f_{i}^{1*}))),  \\
f_i^{2} &= [f_{i, cls}^{1}, f_{i}^{2*}].
\end{aligned}
\end{equation}
The resulting $f_i^{2}$ is then fed into the following ViT blocks after residual connections. Motivated by \cite{chu2021conditional}, through generating convolution kernels dynamically according to class tokens, we also involve class token in depth-wise convolution. In particular, image tokens ${f} \in \mathop{\mathbb{R}}^{(T_{h} \times T_{w}) \times D}$ are transformed to the shape of $\mathop{\mathbb{R}}^ {D \times T_{h} \times T_{w}}$ before convolution. The class token ${X} \in \mathop{\mathbb{R}}^{1 \times D}$ is used to generate a weight matrix $W = \{w_1, w_2, \cdots, w_N\} $ with a fully-connected layer. It is further used to produce a conditional kernel weight as:

\begin{equation}
\begin{aligned}
\mathcal{W}(X) = w_1 \cdot K_1 + \cdots + w_N \cdot K_N,
\end{aligned}
\label{eq:condconv}
\end{equation}
where $K_{i}$ indicates $i$-th learnable kernel which shares same shape with original DWConv. Consequently, we proceed to obtain the final conditional kernel weight $\mathcal{W}(X)$.

\begin{table}[htbp]
  \centering
  \caption{Results of models with both OPE. and convolution in FFN. FFN$_{\text{conv}}$ refers to FFN injected with DWConv, while FFN$_{\text{conv*}}$  indicates conditional DWConv with $N = 8$. }
  \vspace{0.05in}
  \label{tab:combine_eval}
  \resizebox{\linewidth}{!}
  {\begin{tabular}{l|lll}
        \toprule
        \multirow{2}{*}{\textbf{Model}} & {\textbf{IN}} & 
        {\textbf{IN-C}}& {\textbf{IN-3DCC}}\\
        ~ & {\textbf{Acc}$\uparrow$} &  
        {\textbf{mCE}$\downarrow$} &  {\textbf{Acc}$\uparrow$}\\
        \cmidrule{1-1} \cmidrule{2-4} 
        DeiT-S & 79.83 &
        54.60 & 57.60\\
        +OPE$_{k=2}$, FFN$_{\text{conv}}$ & 79.99 & 
        52.29 & \textbf{58.89}\\
        +OPE$_{k=2}$, FFN$_{\text{conv*}}$ & 79.79 & 
        \textbf{52.19} \Drop{2.41} & 58.84 \Rise{1.29} \\
        \midrule
        Swin-T & 81.16 & 
        61.96 & 55.90\\
        +OPE$_{k=4}$, FFN$_{\text{conv}}$ & \textbf{81.98} & 
        \textbf{54.10} \Drop{7.86} & \textbf{58.46} \Rise{2.56} \\
\bottomrule
\end{tabular}}
\end{table}

As Table \ref{tab:ffn_eval} shows, \textit{{adding convolution to FFN brings solid gains on robustness}}. However, replacing with conditional convolution leads to no further gain in robustness both on DeiT-S and Swin-T. Nevertheless, the results are still better than those without convolution, highlighting the potential contributions of convolutional FFN to robustness.

\begin{table*}[htbp]
  \centering
  \caption{We conduct extensive experiments of adversarial noise training on ViTs. Models of DeiT-S and PiT-S are initialized with weights from official open source.  $\dagger$ indicates adding opverlapping patch embedding and convolutional feed-forward network to DeiT-S and $\ddagger$ represents DeiT-S with opverlapping patch embedding and conditional convolutional feed-forward network, as demonstrated in Table \ref{tab:combine_eval}.
  Methods start with Cond refer to ones strengthened with our conditional augmentation strategy. All experiments are conducted with norm setting to 80.}
  \label{tab:noise_eval}

%
{\begin{tabular}{l| p{1.5cm} p{1.8cm} p{1.8cm} l p{1.8cm}}
    \toprule
    \multirow{2}{*}{\textbf{Method}} &
    {\textbf{IN}} & {\textbf{IN-C}}& {\textbf{IN-C}}& {\textbf{IN-C}}  & {\textbf{IN-3DCC}} \\
    ~ & {\textbf{Acc}$\uparrow$} &  {\textbf{Acc}$\uparrow$} &  {\textbf{mCE}$\downarrow$} &  {\textbf{w/o Noise Acc}$\uparrow$} & {\textbf{Acc}$\uparrow$}\\
    
    \cmidrule{1-1} \cmidrule{2-6} 
    DeiT-S~\cite{touvron2021training} & 79.83 & 57.20 & 54.60 &57.05 & 57.60\\
    + finetune 300 ep. & 81.30 & 59.06 & 52.24 & 58.85 & 58.91\\
    \midrule[0.2pt]
    + ANT Gaussian 1$\times$1 & 80.27 & 62.51 & 48.09 & 60.71 & 61.12 \\
    + ANT Gaussian 3$\times$3 & 80.52 & 62.42 & 48.18 & \textbf{62.36} & 61.21\\
    + ANT Gaussian 16$\times$16 & 80.40 & 62.30 & 48.35 & 60.44 & 61.21\\
    + ANT Speckle 1$\times$1 & 80.35 & 62.74 & 47.82 & 60.81 & 61.10 \\
    + ANT Speckle 3$\times$3 & 79.80 &  61.60 & 49.25 & 60.04 & 60.49 \\
    + Cond ANT Gaussian 1$\times$1 &  80.18 & 63.16 &  47.28 & 61.24 &  \textbf{61.38}\\
    + Cond ANT Speckle 1$\times$1 &  80.15 & \textbf{63.28} \Rise{6.08} & \textbf{47.18} \Drop{7.42} & 61.02 \Rise{3.97} & 61.10 \Rise{3.50} \\
    \midrule[0.8pt]
     PiT-S~\cite{heo2021rethinking} & 80.98 & 58.88 & 52.47 & 58.38 & 58.27 \\
     + ANT Gaussian 1$\times$1 & 80.77 &  62.86 & 47.60 &  61.09 & 61.60\\
     + Cond ANT Gaussian 1$\times$1 & 80.80 & 63.13 & 47.28 & 61.52 & 61.38 \\
     + Cond ANT Speckle 1$\times$1 &  80.74 & \textbf{63.64} \Rise{3.76} & \textbf{46.63} \Drop{5.84}& \textbf{62.05} \Rise{3.77}&  \textbf{61.89} \Rise{3.62}\\
    \midrule[0.8pt]
    DeiT-S $\dagger$ & 79.99 & 59.02 & 52.29 & 58.65 & 58.89\\
    + Cond ANT Speckle 1$\times$1 & 79.91 & \textbf{64.73} \Rise{5.71} & \textbf{45.33} \Drop{6.96} &  \textbf{62.75} \Rise{4.10} & \textbf{62.22} \Rise{3.33} \\
    DeiT-S $\ddagger$ & 79.79 & 59.11 & 52.19 & 58.63 & 58.84\\
    + Cond ANT Speckle 1$\times$1 & 79.85 & \textbf{64.16} \Rise{5.05} &  \textbf{46.06} \Drop{6.13} & \textbf{62.13} \Rise{3.50}& \textbf{61.88} \Rise{3.04}\\
     
\bottomrule
\end{tabular}}
\end{table*}

Despite the fact that both adding overlapping patch embedding and injecting convolution into FFN are fairly simple, they demonstrate good effectiveness. Moreover, we incorporate two designs simutaneously into DeiT-S and Swin-T. As displayed in Table \ref{tab:combine_eval}, combined designs achieve further improvement. Ultimately, DeiT-S drops on mCE of ImageNet-C by \textbf{2.41} and Swin-T drops by \textbf{7.86}.

\section{Augmentation Strategies}
As mentioned in related work, various augmentation techniques have been proposed and proven to be effective on CNNs for robustness. To boost robustness of ViTs towards common corruptions, it is natural to build them upon successful CNN-based augmentation methods. However, since ViTs train along with strong augmengtations \eg Mixup~\cite{zhang2017mixup}, CutMix~\cite{yun2019cutmix}, ColorJitter and AutoAugment~\cite{cubuk2018autoaugment}, whether previous 
augmentation still takes effect is uncertain. Among recent work concerning the robustness of CNNs, we lay our emphasis on Adversarial Noise Training (ANT) \cite{rusak2020simple} and fourier-based augmentation \cite{chen2021amplitude} and aim to evaluate their effectiveness on robustness on ViTs.

\subsection{Adversarial Noise Training}
\label{sec:sec_ant} 

ANT~\cite{rusak2020simple} brings great advances on robustness towards common corruptions on CNNs. Random noise is first tuned by a light-weight neural network and cast on images. Afterwards models are trained against enhanced noise adversarially. To specify, in the first place, gaussian (Eq.~\ref{eq:guassion}) noise and speckle noise (Eq.~\ref{eq:speckle}) are generated to perturb image $x$,
\begin{subequations}
    \begin{align}
    \mathbf{\Sigma}_1(x) &= x + \mathcal{C}_{p}(\sigma_{\delta}, \epsilon) ,\label{eq:guassion} \\
    \mathbf{\Sigma}_2(x) &= x + \mathcal{C}_{p}(\sigma_{\delta} \cdot x, \epsilon), \label{eq:speckle} 
    \end{align}
\end{subequations}
among which $\sigma_{\delta} \in \mathbb{R}^{C \times H \times W}$ obeys the distribution of $ \mathcal{N}(0, \delta^2)$. The clipping function $\mathcal{C}_{p}$ confines noise within boundary of $\epsilon$ for $L_{p}$ norm. We conduct ANT on ViT (denoted as $f_{\theta}$) with the following optimization objective:
\vspace{0.05in}
\begin{equation*}
\begin{aligned}
\min_{\theta} \max_{\tau}  \mathop{\mathbb{E}}\limits_{(x,y)\sim D}  \mathop{\mathbb{E}}\limits_{\sigma_{\delta} \sim \mathcal{N}(\mathbf{0}, \delta^2)} & [\mathcal{L} ( f_{\theta}(x + \mathcal{C}_{p}(  \mathcal{P}_1(\sigma_{\delta}), \epsilon), y)], \\
\min_{\theta} \max_{\tau}  \mathop{\mathbb{E}}\limits_{(x,y)\sim D}  \mathop{\mathbb{E}}\limits_{\sigma_{\delta} \sim \mathcal{N}(\mathbf{0}, \delta^2)} & [\mathcal{L} ( f_{\theta}(x + \mathcal{C}_{p}(\mathcal{P}_2(\sigma_{\delta} \cdot x), \epsilon), y)],
\end{aligned}
\end{equation*}
where $\mathcal{P}$ denotes noise generators which typically consist of a 4-layer convolution with $1\times1$ kernel. In the inner loop, the generator is first enhanced to produce most harmful noise, serving as attacks in a similar spirit to adversarial training. In outer loop of defense phase, resistance of ViTs is optimized by learning from noisy inputs.

\vspace{0.1in}
\noindent\textbf{Adversarial noise training is effective} 
For simplicity, we attack ViTs with the ANT generators and defend on noisy data once per iteration. During the attack process, all data are assigned with noise generated by ANT while 50\% of noisy data together with 50\% of clean data is used in the defense phase. In addition, in line with experience replay setting in original ANT, we restart generators with new parameters in each epoch with 20\% of samples augmented by randomly-drawn previous noise generators. We train noise generators with an Adam optimizer and set its learning rate to $8e^{-5}$. All augmentation strategies in DeiT training~\cite{touvron2021training} are kept since they turn out to be beneficial to final results  (\emph{c.f.} Table \ref{tab:ant_aug}).

From Table \ref{tab:noise_eval}, we observe that the ANT is effective on ViTs. DeiT-S with ANT drops 6.51 in mCE of ImageNet-C and mCE of PiT-S decreases by 4.87. In accordance with findings on CNNs, ANT also gives an impetus to robustness against corruptions besides noise. 

\begin{table}[htbp]
  \centering
  \caption{Ablations on sequence of augmentation in Gaussian ANT 1x1 training on pre-trained DeiT-S, where Mix. is the abbreviation of Mixup and CutMix. In the following, our default experimental setting is same with (3).}
  \vspace{-0.05in}
  {\begin{tabular}{l|ccc}
        \toprule
        {\textbf{Augmentation}} & {\textbf{IN}} & \multicolumn{2}{c}{\textbf{IN-C}} \\
        {\textbf{Sequence}} & {\textbf{Acc}$\uparrow$} &  {\textbf{Acc}$\uparrow$} &  {\textbf{mCE}$\downarrow$}\\
        \cmidrule{1-1} \cmidrule{2-4} 
        (1) Early Mix & 80.76 & 60.45 & 50.56 \\
        (2) Full Mix & 80.19 & 61.95 & 48.66 \\
        (3) Separate Mix & 80.27 & \textbf{62.51} & \textbf{48.09} \\
        \bottomrule
        \end{tabular}}
\label{tab:mixup_ant}
\end{table}

\vspace{0.1in}
\noindent\textbf{Basic augmentation is necessary} Since we train ANT on top of the recipe of DeiT, four basic augmentation approaches, \ie, Mixup~\cite{zhang2017mixup}, CutMix~\cite{yun2019cutmix}, ColorJitter and AutoAugment~\cite{cubuk2018autoaugment} are involved so as to stabilize training and boost clean accuracy. There is a necessity to explore interplay between additive adversarial noise and previously-adopted augmentation. Results in Table \ref{tab:ant_aug} indicate that {\textit{ViTs are highly dependent on basic perturbations}}. Absence of any basic augmentation gives rise to performance drops on either ImageNet or ImageNet-C.

\begin{table}[htbp]
  \centering
  \caption{Ablations on exempting basic augmentation from Gaussian ANT 1x1 training on pre-trained DeiT-S. The baseline includes all augmentation. In each run, we remove one type of augmentation.}
    \vspace{-0.05in}
\label{tab:ant_aug}
{\begin{tabular}{l|ccc}
        \toprule
        \multirow{2}{*}{\textbf{Augmentation}} & {\textbf{IN}} & \multicolumn{2}{c}{\textbf{IN-C}} \\
        ~ & {\textbf{Acc}$\uparrow$} &  {\textbf{Acc}$\uparrow$} &  {\textbf{mCE}$\downarrow$}\\
        \cmidrule{1-1} \cmidrule{2-4} 
        ANT 1x1 & 80.27 & \textbf{62.51} & \textbf{48.09} \\
        w/o Mixup &  80.65 & 59.99 & 51.07 \\
        w/o CutMix & 79.94 & 62.13 & 48.62 \\
        w/o ColorJitter & 80.41 & 62.12 & 48.53 \\
        w/o AutoAugment & 79.68 & 58.75 & 52.89 \\
    \bottomrule
    \end{tabular}}
\end{table}

Furthermore, we argue that sequence of Mixup, CutMix and noise augmentation is also crucial to final results. We make the investigation with 3 different strategies: (1) Early Mix: we first implement normalization on data, followed up with Mixup or CutMix and noise is added at last; (2) Full Mix: noise is first applied followed by normalization and Mixup or CutMix is fulfilled along the whole batch; (3) Separate Mix: noise is first applied followed by normalization while Mixup or CutMix is conducted separately among clean and noisy data. Separate mix offers the best results in view of results in Table \ref{tab:mixup_ant}.

\vspace{0.1in}
\noindent\textbf{Structural noise is more powerful} We build adversarial generators both on gaussian and speckle noise. Speckle noise amplified by generators is applied on $1 \times 1$ pixel region and thus bear most structural information of original images. 
As it can be seen from Table \ref{tab:noise_eval}, $1 \times 1$ speckle noise offers the best performance on ImageNet-C. However, $3 \times 3$ convolutional kernel may impair its spatial information since it leads to a lower result than $1 \times 1$ counterpart.

\begin{table}[htbp]
  \centering
  \caption{Results of masked adversarial noise training on DeiT-S. $\dagger$ indicates training from scratch. All experiments are conducted with noise generated by ANT Gaussian 1x1.}
  \vspace{-0.05in}
  {\begin{tabular}{l|ccc}
        \toprule
        \multirow{2}{*}{\textbf{Method}} & {\textbf{IN}} & \multicolumn{2}{c}{\textbf{IN-C}} \\
        ~ & {\textbf{Acc}$\uparrow$} &  {\textbf{Acc}$\uparrow$} &  {\textbf{mCE}$\downarrow$}\\
        \cmidrule{1-1} \cmidrule{2-4} 
        + 50\% mask $\dagger$ & 79.28 & 57.05 & 54.81 \\
        + 30\% mask & 80.90 & 59.89 & 51.41 \\
        + 50\% mask & 80.63 & 59.42 & 51.99 \\
        + 70\% mask & 80.58 & 59.99 & 51.26 \\
    \bottomrule
    \end{tabular}}
\label{tab:noise_mask}
\end{table}

\vspace{0.1in}
\noindent\textbf{Training with masks shows no merit} Patch-wise augmentation has been adopted in recent work \cite{ge2021robust,mao2021towards} and masked image modeling \cite{he2021masked,xie2021simmim} becomes a new trend on ViTs. To take a closer look at characteristics of ViTs, we employ ANT in a masked manner. Instead of training on 50\% noisy images, we impose adversarial gaussian noise on 50\% patches of a single image selected by a random mask. Results in Table \ref{tab:noise_mask} show that noisy-masked training on pre-trained models fails to generalize well on corrupted data while achieving better clean performance. The phenomenon is in line with discovery in \cite{naseer2021intriguing}, where ViTs are reported to be robust against occlusions. In essence, ViTs rely little on corrupted patches and succeed to perceive semantic information from the remaining clean patches.

\begin{table}[htbp]
  \centering
   \caption{Results of adopting amplitude interpolation on DeiT-S. $\dagger$ indicates training from scratch and $p$ denotes portion of data augmented with amplitude interpolation.}
     \vspace{-0.05in}
  \resizebox{0.7\linewidth}{!}{\begin{tabular}{l|ccc}
        \toprule
        \multirow{2}{*}{\textbf{Method}} & {\textbf{IN}} & \multicolumn{2}{c}{\textbf{IN-C}} \\
        ~ & {\textbf{Acc}$\uparrow$} &  {\textbf{Acc}$\uparrow$} &  {\textbf{mCE}$\downarrow$}\\
        \cmidrule{1-1} \cmidrule{2-4} 
        DeiT-S  & 79.83 & 57.20 & 54.60 \\
        + APR$^{*}_{p = 0.5}$ & 80.16 & 59.91 & 51.19 \\
        + APR$^{*}_{p = 0.5}\dagger$ & 74.97 & 51.67 & 61.84 \\
    \bottomrule
    \end{tabular}}
\label{tab:fact_aug}
\end{table}

\subsubsection{Augmentation in Fourier Domain}
According to \cite{yin2019fourier}, adversarial augmentation is blind to corruptions in the low-frequency domain, \ie, Fog and Contrast. ANT also induces a slight drop in corruption categories of Fog on ViTs. Therefore, We perform an augmentation in the frequency domain, which is similar to APR~\cite{chen2021amplitude}, a method effective to promote robustness towards low-frequency corruptions on CNNs by swapping amplitude components between samples. Besides, impressed by FACT~\cite{xu2021fourier} that improves models on domain generalization tasks by linearly interpolating between the amplitude spectrums of two images, we equip APR with interpolation. For samples $x$ and $x'$ chosen randomly within same batch, our method can be denoted as
\begin{equation}
\begin{aligned}
\tilde{\mathcal{A}}(x) &= (1 - \lambda)\mathcal{A}(x) + \lambda \mathcal{A}(x'), \\
\mathcal{F}(x) &= \tilde{\mathcal{A}}(x) \cdot e^{-j \cdot\mathcal{P}(x)},
\label{eq:fact_eq} 
\end{aligned}
\end{equation}
where $\mathcal{A}(x)$ denotes the amplitude component of $x$ and $\mathcal{P}(x)$ refers to the phase component, $\lambda$ is obtained by randomly drawn from $\mathcal{N}(\mathbf{0}, 0.5)$.

\vspace{0.05in}
\noindent\textbf{Amplitude interpolation is inferior on ViTs} Unlike CNNs, training from scratch with fourier-based augmentation induces performance drops in both robustness and clean accuracy, as shown in Table \ref{tab:fact_aug}. To relieve influence of basic augmentation, we also try training without Mixup or CutMix. However, it undergoes a failure of stable training, which also suggests that ViTs heavily rely on basic augmentation. Therefore we switch to training on pre-trained models, resulting in lift in accuracy by 2.71\% in ImageNet-C, however, which is inferior to ANT and having a small gap with direct finetuning in Table \ref{tab:noise_eval}. 

\begin{figure}[htbp]
\centering
\includegraphics[width=\linewidth]{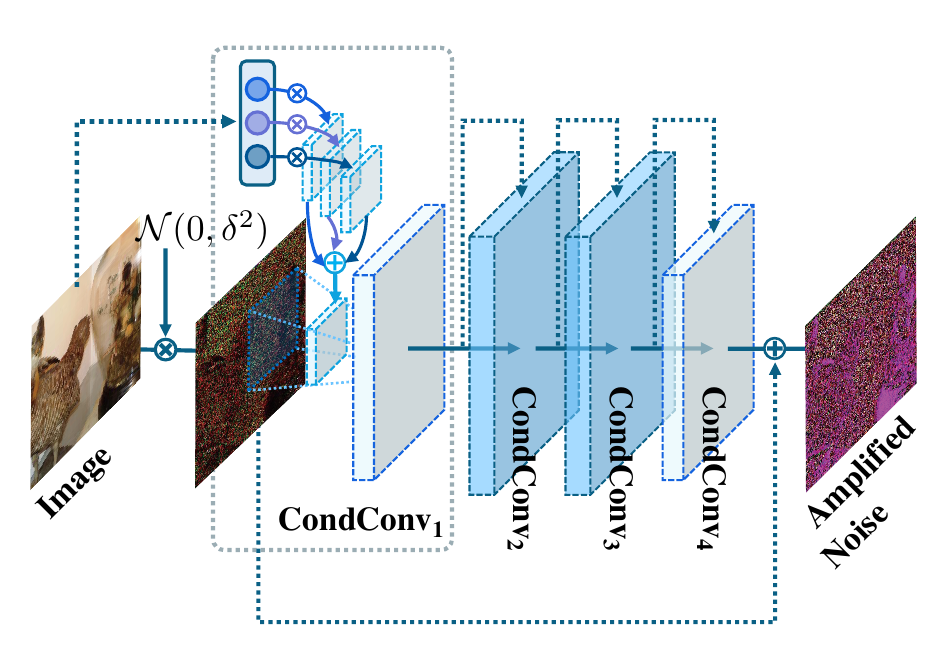}
 \vspace{-0.2in}
\caption{An overview of Conditional ANT framework. Given a clean image and random gaussian noise, speckle noise is acquired by multiplication. Both clean image and noise will be fed into conditional ANT composed of 4 layers of $1\times1$ CondConv and a residual connection. 
}
\label{fig:cond_ant}
\end{figure}

\subsection{Conditional Adversarial Augmentation} 
While ANT is effective, it produces noise regardless of the visual content of input samples with ``one-size-fits-all'' parameters. Recently, there is a growing interest in dynamic networks in CNNs that generate sample-specific parameters conditioned on inputs~\cite{chen2020dynamic,yang2019condconv,zhang2020dynet}. This motivates us to explore dynamic techniques in ANT for improved performance on ViTs. We introduce input-dependent augmentation on top of dynamic networks, which instantiates conditional computation for augmentation  by dynamically generating augmentation parameters for each input instance.

\begin{table}[htbp]
  \centering
  \caption{Comparison with state-of-the-art ViTs for robustness towards common corruption. Cond. ANT indicates conditional ant with speckle noise and $1 \times 1$ kernel. $\dagger$ denotes strengthened noise with norm of 130.}
  \label{tab:sota_cmp}
  \resizebox{\linewidth}{!}
 {\begin{tabular}{l| lll}
    \toprule
    \multirow{2}{*}{\textbf{Model}} &
    {\textbf{IN}} & {\textbf{IN-C}} & {\textbf{IN-3DCC}} \\
    ~ & {\textbf{Acc}$\uparrow$} &  {\textbf{mCE}$\downarrow$} & {\textbf{Acc}$\uparrow$} \\


    \midrule
    RVT-B & \textbf{82.60} & 46.80 & 63.32 \\
    Discrete ViT & 79.48 & 46.22 & - \\
    Adv. Pyramid ViT & 81.71 &  44.99  & - \\
    \midrule
    DeiT-B & 81.80 & 48.52 & 61.45 \\
    DeiT-B + Cond. ANT & 82.19 & 42.41 & 64.42 \\
    \textbf{DeiT-B + Cond. ANT} $\dagger$ & 82.35 \Rise{0.55} & \textbf{40.68} \Drop{7.84} & \textbf{65.31}  \Rise{3.86}\\
\bottomrule
\end{tabular}
}
\end{table}

\vspace{0.1in}
\noindent\textbf{Conditional ANT} 
We wish to generate sample-specific noise conditioned on input images. In practice, we aim to apply dynamic augmentation weights on top of ANT. To this end, we resort to CondConv~\cite{yang2019condconv} which replaces vanilla convolution layers with conditional convolutions in noise generators. Specifically, image $x$ will be mapped to a space of $N$ expert weights by a routing function $\mathcal{R}:  \mathbb{R}^{C \times H \times W} \rightarrow  \mathbb{R}^{N}$. Consequently, a customized kernel $\mathcal{W}$ will be generated by the image-specific weights in the same way as conditional DWConv (Eq. \ref{eq:condconv}). At present we stick to the original routing function in CondConv, which can be elaborated as:
\begin{equation}
\mathcal{R}(x) = \texttt{Sigmoid}(\texttt{FC}(\texttt{GlobalAvgPool}(x))).
\end{equation}

Next, we proceed to fine-tune on ViTs via conditional ANT with $N$ set to 16. Results of experiments suggest that \textit{the dynamic techniques achieves state-of-the-art performance}. As demonstrated in Table \ref{tab:noise_eval}, training DeiT-S with conditional ANT surpasses original ANT by 0.65\% (gaussian noise) and 0.54\% (speckle noise) for Top-1 accuracy on ImageNet-C. On PiT-S, conditional ANT with speckle noise brings about rise of 4.76\% on ImageNet-C accuracy. In addition, by applying conditional ANT to DeiT-S equipped with improved architecture, \ie overlapping patch embedding and convolutional feed-forward network, we achieve 45.33 for mCE on ImageNet-C, which is 4.07 better than RVT-S~\cite{mao2021towards}. 

Furthermore, we conduct conditional augmentation strategy on DeiT-B. Although a trade-off between performance on clean and corrupted data is demonstrated on DeiT-S, applying conditional ANT on DeiT-B conduces to growth on ImageNet by 0.55\%. We owe this result to its larger capacity and hypothesize that heavy augmentation may serve as the catalyst for better performance of large ViTs. Compared with other state-of-the-art ViTs referred in Table \ref{tab:sota_cmp}, DeiT-B with conditional ANT outstands in performance on both ImageNet-C and ImageNet-3DCC and is also highly competitive on clean ImageNet recognition.

\section{Conclusion}
In this work, we provided deeper insights into the robustness of ViTs towards common corruptions by walking through impact of architectural designs and transferability of previous data augmentation methods to ViTs, among which we verified that ViTs benefit from overlapping patch embedding, convolutional feed-forward network and adversarial noise training. We also proposed a novel conditional augmentation strategy which utilizes instance-dependent augmentation parameters to further improve the robustness of ViTs. Comprehensive experiments were conducted and thorough investigations were delivered above.

\bibliography{main}

\appendix


\begin{table}[t!]
  \caption{A thorough analysis of ViTs robustness. Weights of ViTs are all acquired from official public resources. \textbf{*} indicates augmenting models with the proposed conditional adversarial speckle noise. $\dagger$ represents injecting overlapping patch embedding and convolutional feed-forward network.} 
  \label{tab:vit_eval_full}
  \centering
  \resizebox{\linewidth}{!}
  {\begin{tabular}{lc c cc cc}
    \toprule
    \multirow{2}{*}{\textbf{Model}} & {\textbf{$\#$Param}} & {\textbf{IN}} & \multicolumn{2}{c}{\textbf{IN-C}} &  {\textbf{IN-3DCC}}  \\
    ~ & \textbf{(M)} & \textbf{Acc} $\uparrow$ & \textbf{Acc}$\uparrow$ & \textbf{mCE}$\downarrow$ & \textbf{Acc}$\uparrow$\\
    \cmidrule{1-2} \cmidrule{3-6}
    T2T-ViT-7 & 4.3 & 71.7 &  39.45 & 77.02 & 43.84 \\
    PiT-T & 4.9 & 72.8 & 45.70 & 69.11 & 47.98\\
    DeiT-T & 5.7 & 72.1 & 44.23 & 71.13 & 47.44\\
    XCiT-T12/16 & 6.7& 77.1 & 51.03 &  62.47 & 52.51\\
    CrossViT-T & 6.9 & 73.4 &  45.70  & 69.35 & 49.30\\
    PVTv1-T & 13.2 & 75.0 & 37.94 & 79.56 & 46.28\\
    RegionViT-T & 13.8 & 80.4  & 53.99  &  58.74 & 55.44 \\
    PVTv2-B1 & 14.0 & 78.7 & 50.81 & 62.65 & 53.27\\
    NesT-T & 17.1 & 81.5 & 55.93 & 56.39 &  57.88\\
    \cmidrule{1-2} \cmidrule{3-6}
    CvT-13 & 20.0 & 81.6 & 55.27 & 57.06 & 56.85 \\
    T2T-ViT-14 & 21.5 &  81.5 & 57.91 & 53.59 & 57.64 \\
    DeiT-S& 22.0 & 79.8 & 57.20 & 54.60 & 57.60 \\
    \textbf{DeiT-S$\dagger$*} & 22.4 & 79.9 & \textbf{64.73} & \textbf{45.33} & \textbf{62.22}\\
    CSWin-T & 22.3 & 82.8 & 60.70 & 50.22 & 60.67\\
    PiT-S &23.5 & 81.0 & 58.88 & 52.47 & 58.27\\
    Twins-PCPVT-S & 24.1 & 81.2 &  56.31 &  55.85 & 57.89\\
    PVTv1-S & 24.5 & 79.8 & 47.89 & 66.89 & 53.94\\
    PVTv2-B2 & 25.4 & 82.0 & 58.92 & 52.56 & 59.06\\
    XCiT-S12/16 & 26.3 & 82.0 &  59.32 & 51.98 & 59.23 \\
    CrossViT-S & 26.7 & 81.0 & 59.23 & 52.03 & 59.28\\
    CrossFormer-T & 27.8 & 81.5 &   57.48 & 54.39 & 58.40\\
    Swin-T & 28.3 & 81.2 & 51.59 & 61.96 & 55.90\\
    RegionViT-S & 30.6 &  82.6 & 58.86  & 52.62  & 59.14\\
    CrossFormer-S & 30.7 & 82.5 &  58.47 & 53.19 & 58.74 \\
    CvT-21 & 31.6 & 82.5 & 58.04  & 53.52 & 59.23\\
    \cmidrule{1-2} \cmidrule{3-6}
    CSWin-S & 34.6 & 83.6	& 62.83 & 47.56 &62.17  \\
    NesT-S  & 38.3 & 83.3 &  60.22 & 50.83 &61.32  \\
    T2T-ViT-19 & 39.2 & 81.9 & 59.38  & 51.75 & 58.96\\
    RegionViT-M & 41.2 & 83.1 & 60.41 & 50.67& 60.12 \\
    Twins-PCPVT-B & 43.8 & 82.7 & 59.64 & 51.63 & 60.44\\
    PVTv1-M & 44.2 & 81.3 & 51.34 & 62.39 & 56.49\\
    Swin-S & 49.6 & 83.2 & 57.16 & 54.92 & 59.56\\
    CrossFormer-B & 52.0 & 83.4 & 61.66 & 49.10 & 61.19\\
    Twins-SVT-B & 56.1 & 83.2 & 60.40 & 50.67 & 60.68\\
     \cmidrule{1-2} \cmidrule{3-6}
    Twins-PCPVT-L  & 60.9 & 83.1 & 60.03  &  51.13 & 60.97\\
    PVTv1-L & 61.4 & 81.7 & 53.33 & 59.86 & 57.80\\
    T2T-ViT-24 & 64.0 & 82.3 & 61.44 & 49.10 & 60.18\\
    NesT-B & 67.7 & 83.8 &  61.10 & 49.84 & 61.74\\
    RegionViT-B  & 72.7 & 83.2 &  60.02 & 51.18 & 60.11\\
    PiT-B & 73.8 & 82.4 & 62.39 & 48.16 & 61.21\\
    CSWin-B & 77.4 & 84.2 & 64.31 & 45.64 & 63.53 \\
    PVTv2-B5 & 82.0 & 83.8 & 64.12 & 45.90 & 62.96\\
    XCiT-M24/16 & 84.4 & 82.7 & 61.26 & 49.50 & 60.84\\
    DeiT-B & 86.6 & 81.8 & 62.06 & 48.52 & 61.45\\
    \textbf{DeiT-B*} & 86.6 & 82.4 & \textbf{68.38} & \textbf{40.68} & \textbf{65.31} \\
    Swin-B & 88.0 & 83.4 & 57.51 & 66.70 & 59.93\\
    CrossFormer-L & 92.0 & 84.0 & 62.37 &  48.26 & 62.00\\
    Twins-SVT-L & 99.2 & 83.7 & 61.67& 49.10 & 61.79\\
    CrossViT-B  & 104.7 & 82.2 & 63.25 & 46.92 & 62.36  \\
    XCiT-L24/16 & 189.1 & 82.9 & 60.93 & 49.85 & 61.27\\
    \bottomrule
\end{tabular}}
\end{table}

\vspace{1.0in}

\section{Appendix}

\vspace{0.1in}
\section{An Overall Robustness Evaluation}
We make a comprehensive investigation of popular ViTs. As shown in Table \ref{tab:vit_eval_full}, we evaluate various ViT backbones on ImageNet-C and ImageNet-3DCC, including DeiT~\cite{touvron2021training}, CrossFormer~\cite{wang2021crossformer}, CrossViT~\cite{chen2021crossvit}, CSWin~\cite{dong2022cswin}, CvT~\cite{wu2021cvt}, NesT~\cite{zhang2021aggregating}, PiT~\cite{heo2021rethinking}, PVTv1~\cite{wang2021pyramid}, PVTv2~\cite{wang2021pvtv2}, RegionViT~\cite{chen2022regionvit}, Swin Transformer~\cite{liu2021swin}, T2T-ViT~\cite{yuan2021tokens}, Twins~\cite{chu2021twins} and XCiT~\cite{ali2021xcit}.

We also equip models with the proposed architecture designs and augmentation strategy. Our models achieve superior robustness both among base models and among small models.

\section{Robustness towards Specific Corruptions}
We investigate the impact of architecture designs and augmentation on specific corruptions. According to Hendrycks et al. ~\cite{hendrycks2018benchmarking} , we divide common corruptions on ImageNet-C into 4 categories, \ie weather, blur, digital and noise. 

As illustrated in Table~\ref{tab:eval_corrup}, DeiT-S with overlapping embedding outperforms in corruptions of Brightness, JPEG Compression, Contrast, Pixelation, Gaussian Noise, Impulse Noise and Shot Noise. Convolutional designs boost accuracy on corruptions from similar categories. In addition, models with combined architecture designs further improve in corruptions of noise category. 

\begin{table}[h!]
  \caption{Results of DeiT-S with different experimental settings. OPE stands for overlapping patch embedding and FFN$_{\text{conv}}$ refers to FFN injected with DWConv, while conv* indicates conditional DWConv. }
  \label{tab:eval_corrup}
  \centering
  \resizebox{\linewidth}{!}{\begin{tabular}{l|cccc}
  \toprule
  \multirow{2}*{\textbf{Model}}  & \multicolumn{4}{c}{\textbf{Weather}} \\
   & Bright.& Fog & Frost & Snow  \\
   \cmidrule{1-1} \cmidrule{2-5} 
  DeiT-S & 44.93&46.01	&46.18	&49.86\\
  + OPE$_{k=1}$ & \textbf{43.31}	&43.27	&46.16	&48.72\\
  + OPE$_{k=2}$ & 43.63	&46.01	&\textbf{45.34}	&49.33\\
  + OPE$_{k=4}$ &44.28	&\textbf{43.18}	&45.47	&50.67\\
  + FFN$_{\text{conv}}$ &43.59	&45.17	&46.34	&\textbf{47.84}	\\
  + FFN$_{\text{conv*}}$ &44.97	&44.61	&47.27	&49.10	 \\
  \midrule
  + OPE$_{k=2}$, FFN$_{\text{conv}}$ & 43.39 & 46.42 &  45.71 & 47.45\\
  + OPE$_{k=2}$, FFN$_{\text{conv*}}$ & 43.99 & 44.11 & 45.91 & 49.04 \\
  \midrule
  \multirow{2}*{\textbf{ }} & \multicolumn{4}{c}{\textbf{Blur}}  \\
  & Defoc. & Glass & Motion & Zoom \\
   \cmidrule{1-1} \cmidrule{2-5} 
  DeiT-S & 61.58	&71.90	&57.86	&71.89\\
  + OPE$_{k=1}$ & 59.65	&71.25	&56.86	&\textbf{69.20}	\\
  + OPE$_{k=2}$ & 60.71	&71.37	&57.88	&71.50\\
  + OPE$_{k=4}$ & 60.92	&71.43	&56.39	&71.41\\
  + FFN$_{\text{conv}}$ &\textbf{59.39}	&\textbf{70.06}	&\textbf{54.25}	&69.37\\
  + FFN$_{\text{conv*}}$ &60.07	&71.22	&56.19	&69.36\\
  \midrule
  + OPE$_{k=2}$, FFN$_{\text{conv}}$  & 59.64 &  70.79 & 55.08 & 70.48\\
  + OPE$_{k=2}$, FFN$_{\text{conv*}}$ & 59.12 &  70.33 &  56.61 & 70.04\\
  \midrule
  \multirow{2}*{\textbf{ }}  & \multicolumn{4}{c}{\textbf{Digital}}\\
  & Contra. & Elast. & JPEG & Pixel.\\
   \cmidrule{1-1} \cmidrule{2-5} 
  DeiT-S & 42.31	&66.58	&60.42	&59.12\\
  + OPE$_{k=1}$ & 40.40	&67.30	&\textbf{56.06}	&\textbf{52.14}	\\
  + OPE$_{k=2}$ & 40.25	&66.93	&57.03	&52.87\\
  + OPE$_{k=4}$ & 40.91	&67.19	&56.84	&52.75\\
  + FFN$_{\text{conv}}$ &40.97	&\textbf{65.66}	&58.53	&50.48\\
  + FFN$_{\text{conv*}}$ &40.04	&66.12	&60.27	&53.05\\
  \midrule
  + OPE$_{k=2}$, FFN$_{\text{conv}}$  &\textbf{39.64} & 65.80 & 57.05 & 51.32 \\
  + OPE$_{k=2}$, FFN$_{\text{conv*}}$ &39.96 & 65.69 & 56.44 & 51.82 \\
  \midrule
  \multirow{2}*{\textbf{ }}  & \multicolumn{3}{c|}{\textbf{Noise}} & \multirow{2}*{\textbf{mCE}} \\
  & Gauss.& Impulse & \multicolumn{1}{c|}{Shot}  & \\
   \cmidrule{1-1} \cmidrule{2-5} 
  DeiT-S & 46.29&46.39& \multicolumn{1}{c|}{47.72}& 54.60\\
  + OPE$_{k=1}$ & 43.60 &43.60 & \multicolumn{1}{c|}{44.53}&52.40 \\
  + OPE$_{k=2}$ & 44.19	&44.03	& \multicolumn{1}{c|}{45.36} &53.10\\
  + OPE$_{k=4}$ &43.98	&43.74	& \multicolumn{1}{c|}{45.00} &52.94\\
  + FFN$_{\text{conv}}$ &44.60 &44.52	&\multicolumn{1}{c|}{45.35} &52.41\\
  + FFN$_{\text{conv*}}$ &45.40	&45.60	& \multicolumn{1}{c|}{45.77} &53.27\\
  \midrule
  + OPE$_{k=2}$, FFN$_{\text{conv}}$ & 43.67 & 43.48 & \multicolumn{1}{c|}{44.44} & 52.29  \\
  + OPE$_{k=2}$, FFN$_{\text{conv*}}$ & \textbf{43.11} & \textbf{42.70} & \multicolumn{1}{c|}{\textbf{44.05}} & 52.19 \\
  \bottomrule
  \end{tabular}}
  \end{table}

\end{document}